\documentclass[preprint,11pt]{elsarticle}
\usepackage{epstopdf}
\usepackage{caption}
\usepackage{setspace}
\usepackage{amssymb}
\usepackage{multirow}
\usepackage{amsthm}
\usepackage{booktabs}
\usepackage{amsmath}
\usepackage{amssymb}
\usepackage{mathrsfs}
\usepackage{longtable}
\usepackage{lscape}
\usepackage{url}
\usepackage{tabularx}
\usepackage{epsfig}
\usepackage{colortbl}
\usepackage{subfigure}
\usepackage{tikz,mathpazo}
\usepackage{graphicx}
\usepackage{natbib}
\usepackage[noend]{algpseudocode}
\usepackage{algorithm}
\usepackage{algorithmicx}

\newtheorem{myDef}{Definition}

\usetikzlibrary{shapes.geometric, arrows}
\biboptions{numbers,sort&compress}

\begin{document}
\begin{frontmatter}
\title{An approach utilizing negation of extended-dimensional vector of disposing mass for ordinal evidences combination in a fuzzy environment}
\author[address1]{Yuanpeng He}
\author[address1]{Fuyuan Xiao \corref{label1}}

\address[address1]{School of Computer and Information Science, Southwest University, Chongqing, 400715, China}
\cortext[label1]{Corresponding author: Fuyuan Xiao, School of
Computer and Information Science, Southwest University, Chongqing,
400715, China. Email address: xiaofuyuan@swu.edu.cn,
doctorxiaofy@hotmail.com.}
\begin{abstract}
How to measure the degree of uncertainty of a given frame of discernment has been a hot topic for years. A lot of meaningful works have provided some effective methods to measure the degree properly. However, a crucial factor, sequence of propositions, is missing in the definition of traditional frame of discernment. In this paper, a detailed definition of ordinal frame of discernment has been provided. Besides, an innovative method utilizing a concept of computer vision to combine the order of propositions and the mass of them is proposed to better manifest relationships between the two important element of the frame of discernment. More than that, a specially designed method covering some powerful tools in indicating the degree of uncertainty of a traditional frame of discernment is also offered to give an indicator of level of uncertainty of an ordinal frame of discernment on the level of vector.
\end{abstract}
\end{frontmatter}
\section{Introduction}

How to measure the degree of uncertainty of information given has been a attractive topic for years. Lots of relevant works have been completed to provide an appropriate solution to extract virtually meaningful indicative indexes from uncertainties. The representatives of them are $D$ number \cite{Deng2019, Kang2019, Xiao2019a}, Dempster-Shafer evidence theory \cite{book, Dempster1967Upper}, $Z$ number \cite{li2020newuncertainty, Zadeh2011, tian2020zslf} and belief function \cite{Wang2019, Zhou2020, Zhang2020}. All of them contribute to alleviate the level of uncertainty of information provided which can be measured by some other powerful tools, like entropy \cite{gao2020pseudopascal, pan2020Probability, DengeXtropy, yanhy2020entropy} and information volume \cite{Deng2020InformationVolume, deng2021fuzzymembershipfunction, Deng2020ScienceChina}. The tools can be regarded as a kind of indicator of reliability of the evidences offered which can be utilized as a reference in combination of evidences. Besides, some other improved methods like optimized rules of combination \cite{https://doi.org/10.1002/int.22366, Liu2020Combination, 1111111}, soft theory \cite{Song2019POWA, fei2019evidence, fei2019intuitionistic} and ordered weight averaging (OWA) operator \cite{Yager2009, Fei2019a, fei2019new} and complex mass function \cite{Xiao2019b, Xiao2020b, Xiao2020e} which solves the problem of management from the level of complex field. 

In general, in the field of traditional process of handling uncertain information, the foundation of theory is satisfying and solid which is also widely applied into different applications like target recognition \cite{LiuF2020TFS, Pan2020, Han2018}, decision making \cite{Han2019, Fei2019b, Song2019a} and pattern classification \cite{8944285, Liu2019, song2018evidence}. However, an important factor is missing when disposing information from different sources, it is the sequence of things to take place. Some works have taken the order into consideration \cite{feng2019lexicographic}. But for the frame of discernment proposed in Dempster-Shafer evidences theory, there exists no related works. Therefore, in this passage, a completely new method is proposed to organically combine sequential information. A crucial formula in the field of computer vision is generalized and specially designed to adapt to manifest features of propositions and the corresponding mass of them to obtain the degree of uncertainty of each evidences utilizing negation theory and the reverse version of the method to get the conflicting parameter. In the last, all of the ordinal evidences are properly and accordingly combined using the indicators harvested in the before.

The rest of the passage are organized as follows. In the section of preliminaries, some relevant concepts are briefly introduced to provide a background for the problem to be solved. The part of proposed method offers details of note of the proposed method and specific process of every step. Some numerical examples and corresponding application are also provided to verify the validity and correctness of the proposed method. In the last, some conclusions are made to summarize the contributions of the theory proposed in this paper.

\section{Preliminaries}
In this section, some relative concepts are roughly introduced. And lots of related works are done according the corresponding theories \cite{Luo2020negation, Fei2019, Deng2019b, yager1987dempster}.
\subsection{Dempster - Shafer evidence theory}
\begin{myDef}(Frame of discernment)\end{myDef}
Let $\Theta$ be a set whose elements are mutually exclusive and collectively exhaustive and the number of elements is $n$. The set is also called as frame of discernment (FOD), which is defined as \cite{book,Dempster1967Upper}:

\begin{equation}
	\Theta = \{N_{1},N_{2},...,N_{i},...,N_{n}\}
\end{equation}

On the base of definition of frame of discernment, the power set of set $\Theta$ which contains $2^{|\Theta|}$ elements is denoted by $2^{\Theta}$, which is defined as:

\begin{equation}
	2^{\Theta} = \{\emptyset,\{N_{1}\},\{N_{2}\},...,\{N_{n}\},...,\{N_{1},N_{2}\},...,\{N_{1},N_{2},...,N_{i}\},...,\Theta\}
\end{equation}
\begin{myDef}(Mass function)\end{myDef}
Assume there exists a frame of discernment $\Theta = \{N_{1},N_{2},...,N_{i},...,N_{n}\}$, a mapping which is also called a mass function, which is defined as:

\begin{equation}
	m: 2^{\Theta} \rightarrow [0,1]
\end{equation}

And the properties it satisfies are defined as:

\begin{equation}
	m(\emptyset) = 0
\end{equation}

\begin{equation}
\sum_{N \in 2^{\Theta}}m(N) = 1
\end{equation}

According to the definition given by Dempster-Shafer evidence theory, a mass function can be also named as a basic probability assignment (BPA). The value distributed to proposition indicate the degree of belief to certain propositions. The bigger the corresponding values of propositions are, the more likely the corresponding incidents to propositions is supposed to take place. In other words, the values assigned to propositions represent the degree of support evidences provide to certain propositions. Besides, a propositions is called a focal element if $m(N) > 0$.

\begin{myDef}(Rule of combination of Dempster-Shafer evidence theory)\end{myDef}
Assume a group of multi-source evidences are provided, in order to get a comprehensive judgement on the actual situation, a rule of combination for evidences given is provided in Dempster-Shafer evidence theory which is defined as:

\begin{equation}
	m(E) = \frac{1}{1 - K}*\sum_{G_{i}\cap H_{j} \cap U_{k} \cap ... = E} m(G)*m(H)*...*m(U)
\end{equation}

where the parameter $K$ indicates the degree of conflict among evidences. The process of calculation of parameter $K$ is defined as:

\begin{equation}
	K = \sum_{G_{i}\cap H_{j} \cap U_{k} \cap ... = \emptyset} m(G)*m(H)*...*m(U)
\end{equation}

\subsection{Intuitionistic environment}
\begin{myDef}(Intuitionistic fuzzy sets)\end{myDef}

Let $R$ be a finite universe of discourse and $B$ be an IFS in the universe, which can be defined as:

\begin{equation}
	R = \{(x,\mu_{R}(x),\upsilon_{R}(x))|x \in B\}
\end{equation}

Where $\mu_{R}(x)$ represents the degree of membership and $\upsilon_{R}(x)$ represents the degree of non-membership. The properties they satisfy are defined as:

\begin{equation}
	\mu_{R}(x): B \rightarrow [0,1]
\end{equation}

\begin{equation}
\upsilon_{R}(x): B \rightarrow [0,1]
\end{equation}

Besides, there exists a constraint on the two parameters of $x \in B$ which is defined as:

\begin{equation}
	0 \leq \mu_{R}(x) + \upsilon_{R}(x) \leq 1
\end{equation}

For an IFS $R$, a degree of hesitance  of $x \in B$ is defined as:

\begin{equation}
	\pi_{R}(x) = 1 - \mu_{R}(x) - \upsilon_{R}(x)
\end{equation}

which reflects the degree of hesitance of the decision system when making a  judgement on the current situation.
\subsection{Negation theory}
\begin{myDef}(Process of negation)\end{myDef}
Assume there exists a frame of discernment and $n$ focal elements $el_{i}$ which are contained in the frame given. Besides, the corresponding value of them is denoted by $m_{Ori}(el_{i})$. The core idea of negation theory is to obtain the reverse side of values provided. The detailed process of obtaining negations is divided into three steps.

\textbf{Step 1: }For every focal element $el_{i}$, the initial value of it is replaced by  $1 - m_{Ori}(el_{i})$ which is regarded as a complementary part of its original mass. The process of obtaining the negation is defined as:

\begin{equation}
el_{i} = m_{Ori}(el_{i}) \rightarrow \widetilde{el_{i}} = 1 - m_{Ori}(el_{i})
\end{equation}

\textbf{Step 2: }The sum of every negation of focal element, $\varpi$, is calculated and the process is defined as:

\begin{equation}
\varpi = \sum_{i}^{n}\widetilde{el_{i}}
\end{equation}

\textbf{Step 3: }It can be easily concluded that the sum $\varpi$ is not equal to 1. Then, the third step is designed to normalize the degree of negation of belief assigned to every proposition which is defined as: 

\begin{equation}
	\widetilde{el_{i}} = 1 - m_{Ori}(el_{i}) \rightarrow \widetilde{el_{i}}^{Nor} = \frac{1 - m_{Ori}(el_{i})}{\varpi}
\end{equation}

At last, a general formula for the final mass after modification corresponding to every proposition can be expressed as:

\begin{equation}
\widetilde{el_{i}}^{Nor} = \frac{1 - m_{Ori}(el_{i})}{n - 1}
\end{equation}

\section{Proposed method}
\subsection{Ordinal frame of discernment}
Compared with the traditional frame of discernment (FOD), for elements contained in the ordinal one (OFOD), the sequence of them matters a lot. With respect to an ordinal set $\Theta_{Ordinal} = \{P_{1}^{1},P_{2}^{2},...,P_{n}^{n}\}$, the subscripts of propositions represent the number of propositions contained in the frame of discernment and the superscripts of propositions represent the sequence of them. Then, all of the properties the elements contained in a frame of discernment are defined as: 

\begin{itemize} 
\item The proposition $P_{1}^{1}$ is expected to be confirmed in the first place. And in the following, the proposition $P_{2}^{2}$ is ascertained and then the rest of the group of propositions is managed in the same manner.
\item The relationship among propositions is only involved with the sequence of propositions. Beyond this, all of the definitions of ordinal frame of discernment remain consistent with the ones given in the traditional Dempster-Shafer evidence theory.
\item With every step of confirming a proposition, the whole situation of the system given is further ascertained.
\end{itemize}

\subsection{The presentation of an OFOD in the form of vector}
Assume there exists a OFOD, which is given as $\Theta_{Ordinal} = \{P_{1}^{1},P_{2}^{2},...,P_{n}^{n}\}$. For a certain proposition, for instance, $P_{1}^{1}$, can be expressed as a form like $\{Order_{P_{1}^{1}}, Mass_{P_{1}^{1}}\}^{T}$. And the rest of the group of propositions can be organized in the same manner. The detailed process of presenting a proposition contained in an OFOD, $Pr$, is defined as:

\begin{equation}
	Pr = \{Order_{Pr}, Mass_{Pr}\}^{T}
\end{equation}

For a complete OFOD, $M$, every element is denoted by $Pr_{i}$, then the whole frame can be defined as:
\begin{equation}
	M = \{\{Order_{Pr_{1}}, Mass_{Pr_{1}}\}^{T},\{Order_{Pr_{2}}, Mass_{Pr_{2}}\}^{T},...,\{Order_{Pr_{n}}, Mass_{Pr_{n}}\}^{T}\}
\end{equation}

Then, it is the vectorial presentation of the OFOD.

\subsection{Modified radial basis function for OFOD}

Radial basis function is utilized in lots of fields, such as computer vision, which plays an important role in different kinds of data processing. The original form of the function is defined as:

\begin{equation}
	RAF = e^{-\sigma||x - y||^{2}}
\end{equation}

where the icon $\sigma$ represents the parameter of width of the function. In order to adapt the function into the field of OFOD, some modifications are made to amend the original function and the amended function is called MRBF which is defined as:

\begin{equation}
	MRBF = e^{-K||Order_{Pr_{1}} - Mass_{Pr_{1}}||^{2}}
\end{equation}

The parameter $\sigma$ is replaced by the measurement of conflict among evidences to better manifest actual conditions of the group of evidences provided. Besides,  in order to have an all-around description on the components of OFOD, parameter $x$ is replaced by the order of propositions $Order_{Pr_{1}}$ and parameter $y$ is replaced by the corresponding mass of propositions $Mass_{Pr_{1}}$, which provides a complete figure on the OFOD.

\subsection{Expanded form of MRBF utilizing Taylor formula}
Taylor formula is a crucial content in advanced mathematics which provides a solution in transferring some complex functions into the form of simple polymerizations. The key role of Taylor formula in simplifying what is complicated is the main reason to make itself a powerful tool in analysing and researching lots of mathematics problems. The details of the formula is defined as:

\begin{equation}
	f(x) = \frac{f(a)}{0!}+\frac{f'(a)}{1!}(x-a)+...+\frac{f^{n}(a)}{n!}(x -a)^{n}+R_{n}(x)
\end{equation}

To disintegrate the formula which represents MRBF, the Taylor formula is utilized to simplify and visualize the information transmitted. All of the provided parameters are consistent with the ones given in the last subsection. Then the procedure of derivation is expressed as:

$exp(-K||Order_{Pr_{i}} - Mass_{Pr_{i}}||^{2})$

$= exp(-K(Order_{Pr_{i}})^{2}) * exp(-K(Mass_{Pr_{i}})^{2}) * exp(K(2 * Order_{Pr_{i}} *  Mass_{Pr_{i}}))$

$=exp(-K(Order_{Pr_{i}})^{2}) * exp(-K(Mass_{Pr_{i}})^{2}) *(\sum_{j =0}^{\infty}\frac{K(2 * Order_{Pr_{i}} *  Mass_{Pr_{i}})^{j}}{j!})$

$=\sum_{i =0}^{\infty}(exp(-K(Order_{Pr_{i}})^{2}) * exp(-K(Mass_{Pr_{i}})^{2})*\sqrt{\frac{2K^{j}}{j!}}*\sqrt{\frac{2K^{j}}{j!}}*(Order_{Pr_{i}})^{j}*(Mass_{Pr_{1}})^{j})$

$=\ell(Order_{Pr_{i}})^{T}\ell(Mass_{Pr_{i}})$

with the infinite dimensional $\ell(Order_{Pr_{i}}) = exp(-K(Order_{Pr_{i}})^{2}) * (1, \\\sqrt{\frac{2}{1!}}Order_{Pr_{i}}, \sqrt{\frac{2^{2}}{2!}}Order_{Pr_{i}}^{2},...)$ and $\ell(Mass_{Pr_{i}}) = exp(-K(Mass_{Pr_{i}})^{2}) * (1, \\\sqrt{\frac{2}{1!}}Mass_{Pr_{i}}, \sqrt{\frac{2^{2}}{2!}}Mass_{Pr_{i}}^{2},...)$. In order to keep enough accuracy in manifesting all features which are owned by corresponding propositions, four dimensions of the data obtained are selected as a reference and adopted into calculation for evaluation for the OFOD. Besides, after confirming the main demand in disposing data provided, the detailed process of obtaining polymerized values of homologous propositions contained in the OFOD can be defined as:

\begin{equation}
	Pr_{1}\rightarrow
	\begin{Bmatrix}
	Order_{Pr_{1} }\\ Mass_{Pr_{1}}
	\end{Bmatrix}\rightarrow
	\begin{Bmatrix}
	e^{-KOrder_{Pr_{1}}^{2}}e^{-KMass_{Pr_{1}}^{2}}\\
	\frac{2^{1}}{1!}Order_{Pr_{1} }Mass_{Pr_{1}}e^{-KOrder_{Pr_{1}}^{2}}e^{-KMass_{Pr_{1}}^{2}}\\
	\frac{2^{2}}{2!}Order_{Pr_{1} }^{2}Mass_{Pr_{1}}^{2}e^{-KOrder_{Pr_{1}}^{2}}e^{-KMass_{Pr_{1}}^{2}}\\
	\frac{2^{3}}{3!}Order_{Pr_{1} }^{3}Mass_{Pr_{1}}^{3}e^{-KOrder_{Pr_{1}}^{2}}e^{-KMass_{Pr_{1}}^{2}}
	\end{Bmatrix}
\end{equation}

The extension of the values of propositions given becomes components of coordinates in four-dimensional space, which achieves a mapping from low dimension to high dimension of data. In order to simplify the process of manifesting every part of the vectorial form of propositions, some simplified icons are designed to represent corresponding complex formulas which are given as:

$M_{1}^{P_{i}} = e^{-KOrder_{Pr_{i}}^{2}}e^{-KMass_{Pr_{i}}^{2}}$,

 $M_{2}^{P_{i}} = \frac{2^{1}}{1!}Order_{Pr_{i} }Mass_{Pr_{i}}e^{-KOrder_{Pr_{i}}^{2}}e^{-KMass_{Pr_{i}}^{2}}$,

$M_{3}^{P_{i}} =\frac{2^{2}}{2!}Order_{Pr_{i} }^{2}Mass_{Pr_{i}}^{2}e^{-KOrder_{Pr_{i}}^{2}}e^{-KMass_{Pr_{i}}^{2}}$,

 $M_{4}^{P_{i}} = \frac{2^{3}}{3!}Order_{Pr_{i} }^{3}Mass_{Pr_{i}}^{3}e^{-KOrder_{Pr_{i}}^{2}}e^{-KMass_{Pr_{i}}^{2}}$.
 
\subsection{Preliminary expanded representation of OFOD}

According to the definition of MRBF, a complete OFOD which is given as $\{Pr_{1},Pr_{2},Pr_{3}\}$ can be expressed as:

\begin{equation}
	\begin{Bmatrix}
	Pr_{1}\\ Pr_{2}\\Pr_{3}
	\end{Bmatrix}\rightarrow
	\begin{Bmatrix}
	M_{1}^{Pr_{1}}&M_{2}^{Pr_{1}}&M_{3}^{Pr_{1}}&M_{4}^{Pr_{1}}\\
	M_{1}^{Pr_{2}}&M_{2}^{Pr_{2}}&M_{3}^{Pr_{2}}&M_{4}^{Pr_{2}}\\
	M_{1}^{Pr_{3}}&M_{2}^{Pr_{3}}&M_{3}^{Pr_{3}}&M_{4}^{Pr_{3}}\\
	\end{Bmatrix}
\end{equation} 

Utilizing a method in enhancing dimensions of data provided, an OFOD can be extended into a form of matrix which has a better ability in presenting relations between values and sequence of propositions. Multiple dimensions offers a completely new point of view in considering OFOD.

\subsection{Structure majorization utilizing intuitionistic environment}

Fuzzy environment has been widely applied into different kinds of fields to solve lots of practical applications, such as decision making and fault diagnosis. Therefore, it is necessary to introduce the basic of fuzzy theory, intuitionistic fuzzy sets, into the whole process of transformation of provided OFOD to improve the accuracy and correctness of decisions made by the whole system. Assume the three propositions provided in the last subsection, $\{Pr_{1},Pr_{2},Pr_{3}\}$, can be also expressed as $\{a,b,\{a,b\}\}$. Then, with respect to propositions offered, the situation of the propositions can be toughly further confirmed as:

$a:$

$Membership_{a}^{Original} = \{M_{1}^{Pr_{1}}, M_{2}^{Pr_{1}}, M_{3}^{Pr_{1}}, M_{4}^{Pr_{1}}\}$

$Non-membership^{Original}_{a} = \{M_{1}^{Pr_{2}}, M_{2}^{Pr_{2}}, M_{3}^{Pr_{2}}, M_{4}^{Pr_{2}}\}$

$Hesitancy^{Original}_{a} = \{	M_{1}^{Pr_{3}}, M_{2}^{Pr_{3}}, M_{3}^{Pr_{3}}, M_{4}^{Pr_{3}}\}$

$b:$

$Membership_{b}^{Original} = \{M_{1}^{Pr_{2}}, M_{2}^{Pr_{2}}, M_{3}^{Pr_{2}}, M_{4}^{Pr_{2}}\}$

$Non-membership_{b}^{Original} =\{ M_{1}^{Pr_{1}}, M_{2}^{Pr_{1}}, M_{3}^{Pr_{1}}, M_{4}^{Pr_{1}}\}$

$Hesitancy_{b}^{Original} = \{M_{1}^{Pr_{3}}, M_{2}^{Pr_{3}}, M_{3}^{Pr_{3}}, M_{4}^{Pr_{3}}\}$

$\{a,b\}:$

$Co_{1}^{Original} = M_{1}^{Pr_{1}} + M_{1}^{Pr_{2}} +M_{1}^{Pr_{3}}$

$Co_{2}^{Original} = M_{2}^{Pr_{1}}+M_{2}^{Pr_{2}}+M_{2}^{Pr_{3}}$

$Co_{3}^{Original} = M_{3}^{Pr_{1}}+M_{3}^{Pr_{2}}+M_{3}^{Pr_{3}}$

$Co_{4}^{Original} = M_{4}^{Pr_{1}}+M_{4}^{Pr_{2}}+M_{4}^{Pr_{3}}$

$Membership_{\{a,b\}}^{Original} =\{Co_{1},Co_{2},Co_{3},Co_{4}\}$

$Non-membership^{Original}_{\{a,b\}} = 0$

$Hesitancy^{Original}_{\{a,b\}} = 0$

However, due to the particularity of modified mass given by MRBF, a step of normalization is designed to standardize vectorial presentations of an offered OFOD. For propositions $a$, the detailed process of normalization can be expressed as:

$vMembership_{a}^{Original} = M_{1}^{Pr_{1}}+ M_{2}^{Pr_{1}}+ M_{3}^{Pr_{1}}+ M_{4}^{Pr_{1}}$,

$vNon-Membership_{a}^{Original} = M_{1}^{Pr_{2}}+ M_{2}^{Pr_{2}}+ M_{3}^{Pr_{2}}+ M_{4}^{Pr_{2}}$,

$vHesitancy_{a}^{Original} = M_{1}^{Pr_{3}}+ M_{2}^{Pr_{3}}+ M_{3}^{Pr_{3}}+ M_{4}^{Pr_{3}}$

$MRatio_{a}^{Final} = \frac{vMembership_{a}^{Original}}{vMembership_{a}^{Original}+vNon-membership^{Original}_{a}+vHesitancy^{Original}_{a}}$

$NRatio^{Final}_{a} = \frac{vNon-membership^{Original}_{a}}{vMembership_{a}^{Original}+vNon-membership^{Original}_{a}+vHesitancy^{Original}_{a}}$

$HRatio^{Final}_{a} = \frac{vHesitancy^{Original}_{a}}{vMembership_{a}^{Original}+vNon-membership^{Original}_{a}+vHesitancy^{Original}_{a}}$

$M_{1}^{Pr_{1_{Final}}} = M_{1}^{Pr_{1}} \times MRatio_{a}^{Final}$, $M_{2}^{Pr_{1_{Final}}} = M_{2}^{Pr_{1}} \times MRatio_{a}^{Final}$,

$M_{3}^{Pr_{1_{Final}}} = M_{3}^{Pr_{1}} \times MRatio_{a}^{Final}$, $M_{4}^{Pr_{1_{Final}}} = M_{4}^{Pr_{1}} \times MRatio_{a}^{Final}$.

$M_{1}^{Pr_{2_{Final}}} = M_{1}^{Pr_{2}} \times MRatio_{a}^{Final}$, $M_{2}^{Pr_{2_{Final}}} = M_{2}^{Pr_{2}} \times MRatio_{a}^{Final}$,

$M_{3}^{Pr_{2_{Final}}} = M_{3}^{Pr_{2}} \times MRatio_{a}^{Final}$, $M_{4}^{Pr_{2_{Final}}} = M_{4}^{Pr_{2}} \times MRatio_{a}^{Final}$.

$M_{1}^{Pr_{3_{Final}}} = M_{1}^{Pr_{3}} \times MRatio_{a}^{Final}$, $M_{2}^{Pr_{3_{Final}}} = M_{2}^{Pr_{3}} \times MRatio_{a}^{Final}$,

$M_{3}^{Pr_{3_{Final}}} = M_{3}^{Pr_{3}} \times MRatio_{a}^{Final}$, $M_{4}^{Pr_{3_{Final}}} = M_{4}^{Pr_{3}} \times MRatio_{a}^{Final}$.

\textbf{Note: }An element is a sum of every component contained in its vectorial presentation. After the step of normalization, every component of the vectorial presentation is expected to be reduced or amplified according to the corresponding proportions obtained through the three typical parameters.

Besides, for the multiple proposition, $\{a,b\}$, the process of obtaining the normalized values of ordinal propositions can be expressed as:

$Ratio_{1}^{Final} = \frac{Co_{1}^{Original}}{Co_{1}^{Original}+Co_{2}^{Original}+Co_{3}^{Original}+Co_{4}^{Original}} $

$Ratio_{2}^{Final} = \frac{Co_{2}^{Original}}{Co_{1}^{Original}+Co_{2}^{Original}+Co_{3}^{Original}+Co_{4}^{Original}} $

$Ratio_{3}^{Final} = \frac{Co_{3}^{Original}}{Co_{1}^{Original}+Co_{2}^{Original}+Co_{3}^{Original}+Co_{4}^{Original}} $

$Ratio_{4}^{Final} = \frac{Co_{4}^{Original}}{Co_{1}^{Original}+Co_{2}^{Original}+Co_{3}^{Original}+Co_{4}^{Original}}$

$Normalization \ \ for\ \ Co_{1}^{Original}:$

$ M_{1}^{Pr_{1_{Final}}} = M_{1}^{Pr_{1}} \times Ratio_{1}^{Final}$,
$ M_{1}^{Pr_{2_{Final}}} = M_{1}^{Pr_{1}} \times Ratio_{1}^{Final}$,

$ M_{1}^{Pr_{3_{Final}}} = M_{1}^{Pr_{1}} \times Ratio_{1}^{Final}$.

$ M_{2}^{Pr_{1_{Final}}} = M_{2}^{Pr_{1}} \times Ratio_{2}^{Final}$,
$ M_{2}^{Pr_{2_{Final}}} = M_{2}^{Pr_{2}} \times Ratio_{2}^{Final}$,

$ M_{2}^{Pr_{3_{Final}}} = M_{2}^{Pr_{3}} \times Ratio_{2}^{Final}$.

$ M_{3}^{Pr_{1_{Final}}} = M_{3}^{Pr_{1}} \times Ratio_{3}^{Final}$,
$ M_{3}^{Pr_{2_{Final}}} = M_{3}^{Pr_{1}} \times Ratio_{3}^{Final}$,

$ M_{3}^{Pr_{3_{Final}}} = M_{3}^{Pr_{1}} \times Ratio_{3}^{Final}$.

For $Co_{2}^{Original}$, $Co_{3}^{Original}$ and $Co_{4}^{Original}$, the step of normalization of them is carried out in the same manner of $Co_{1}^{Original}$. And all of the values are restricted into a normative field. 

\textbf{Note: }Although the extended form of every separate element is summed, every dimension is still regarded as an individual instead of a whole.

\subsection{Negation for expanded vectorial OFOD}
According to the original definition of negation theory, an improved method of negation for vectorial OFOD is designed to manifest the features the provided OFOD has on the opposite side. And assume there exists an OFOD which is exactly the same as the one offered in last subsection and the process for proposition $a$, is defined as:

\begin{equation}
	a \rightarrow 
	\begin{Bmatrix}
Membership_{a}\\ Non-membership_{a}\\Hesitancy_{a}
	\end{Bmatrix} \rightarrow
	\begin{Bmatrix}
		M_{1}^{a}& M_{2}^{a}& M_{3}^{a}& M_{4}^{a}\\
	M_{1}^{b}& M_{2}^{b}& M_{3}^{b}& M_{4}^{b}\\
	M_{1}^{\{a,b\}}& M_{2}^{\{a,b\}}& M_{3}^{\{a,b\}}& M_{4}^{\{a,b\}}\\
	\end{Bmatrix}
\end{equation}

Because every component of the matrix is normalized, the values of them is expected to be less than $1$. In this step, the operation is using the number $1$ to minus each of the component and the results of $1 - M_{i}^{Pr_{i}}$ is denoted by $G_{i}^{Pr_{i}}$. Besides, the process of the operation is defined as:

\begin{equation}
		\begin{Bmatrix}
	M_{1}^{a}& M_{2}^{a}& M_{3}^{a}& M_{4}^{a}\\
	M_{1}^{b}& M_{2}^{b}& M_{3}^{b}& M_{4}^{b}\\
	M_{1}^{\{a,b\}}& M_{2}^{\{a,b\}}& M_{3}^{\{a,b\}}& M_{4}^{\{a,b\}}\\
	\end{Bmatrix} \rightarrow 
	\begin{Bmatrix}
	G_{1}^{a}& G_{2}^{a}& G_{3}^{a}& G_{4}^{a}\\
	G_{1}^{b}& G_{2}^{b}& G_{3}^{b}& G_{4}^{b}\\
	G_{1}^{\{a,b\}}& G_{2}^{\{a,b\}}& G_{3}^{\{a,b\}}& G_{4}^{\{a,b\}}\\
	\end{Bmatrix}
\end{equation}

However, after the process of negation, the values of each of the component is not under any reasonable limits. Therefore, another step of normalization is designed to put every of them into a structured rule. And the detailed process of the step of normalization is defined as:

\begin{equation}
	G_{i}^{Pr_{j}^{Final}} = \frac{G_{i}^{Pr_{i}}}{\sum_{i = 1}^{n}\sum_{j = 1}^{m}	G_{i}^{Pr_{j}}}
\end{equation}

\textbf{Note : }The value of parameter $n$ and $m$ is fluctuated according to the situation of OFOD and the role of them is to ensure all the values of components contained in the matrix are taken into consideration.

And after the step of negation, the information contained in the matrix is all about the opposite of the offered OFOD which is more useful in  indicating some underlying relations of it compared with evaluating the frame by utilizing its original mass provided. Besides, the presentation in the form of matrix is called MOOFOD.

\subsection{Proposed method to measure the degree of uncertainty of OFOD}

In the original definition of Dempster-Shafer evidence theory, the degree of conflict among evidences is represented by a parameter $K$. On the opposite, if the values which are taken into calculation is the negation of them, then the rules for calculation is expected to be altered correspondingly. Besides, considering different kinds of factors, the process of obtaining the degree of uncertainty is defined as:

\begin{equation}
	U(Pr_{j}) = \sum_{y \neq i}M_{y}^{Pr_{j}^{Final}} G_{i}^{Pr_{j}^{Final}}+\sum_{y \neq i}\sum_{j \neq f}M_{y}^{Pr_{j}^{Final}} G_{i}^{Pr_{f}^{Final}}
\end{equation}

\textbf{Note: }$M_{y}^{Pr_{j}^{Final}}$ and $G_{i}^{Pr_{f}^{Final}}$ do not lie in the same environment. For example, $M_{1}^{Pr_{1}^{Final}}$ is a membership and $G_{2}^{Pr_{2}^{Final}}$ is a non-membership which is allowed. But $M_{1}^{Pr_{1}^{Final}}$ and $G_{2}^{Pr_{2}^{Final}}$ can not be memberships or non-membership or hesitancy at the same time. And for the propositions of same dimension, it is regarded as the same position like traditional frame of discernment.

According to the definition of the conflicting parameter $K$ and considering the difference among separate kinds of provided evidences, with respect to propositions $Pr_{j}$, the formula to obtain the degree not only indicate the difference between original mass and the negation of other dimensions of it, but also present conflicts among different piece of evidences in different dimensions. However, because the values which are taken into calculation, the standard of judgements is altered correspondingly. Therefore, it can be easily concluded that the smaller the values obtained are, the bigger the degree of uncertainty of the specific evidence is. Then, in this step, the measurement of the degree of uncertainty of the evidence is completed.

\subsection{Rule of combination of OFOD in vectorial presentation}

Dempster-Shafer evidence theory provides a reasonable method to combine different evidences from separate information sources. However, the rule of combination dose not takes the influences brought by sequences of propositions into consideration. In this paper, a extended-dimensional method to combine sequence and mass of propositions organically is designed to manage relationships of propositions contained in an OFOD. In this section, a specially customised method for combination in OFOD is provided to combine ordinal propositions from different sources guaranteeing correctness and validity of the results obtained. The detailed process of combination can expressed as follows.

The weights for every evidence can be defined as:

\begin{equation}
	W^{Pr_{j}}_{t} = \frac{U(Pr_{j})}{\sum_{i = 1}^{n}U(Pr_{i})}
\end{equation}

Then, if there exists more than one piece for the proposition, the modified value for one single proposition in one of dimensions is defined as:

\begin{equation}
	M_{y_{Final}}^{Pr_{j}^{Final}} = \frac{\sum W^{Pr_{j}}_{t_{Evidence_{w}}} \times M_{y_{Evidence_{w}}}^{Pr_{j}^{Final}}}{\sum_{i = 1}^{n}\sum W^{Pr_{j}}_{t_{Evidence_{i}}} \times M_{y_{Evidence_{i}}}^{Pr_{j}^{Final}}}
\end{equation}

All in all, the obtained values are optimized results for propositions contained in an OFOD which provides an intuitive and accurate judgements on practical situations considering sequence of propositions. 

In the last, the rule of combination of Dempster-Shafer evidence theory is utilized $n-1$ times to acquire the final mass of indicator of specific incidents. 

\section{Validation of the proposed method}
\subsection{Numerical examples}

In this section, some meaningful examples are provided to verify the validity and correctness of the proposed method.

\textbf{Example 1: }

Assume there exists an OFOD which contains four propositions $\{a,b,c,\{a,b\}\}$ and the values of the propositions are listed in the Table \ref{3}. Besides, the extended-dimensional form of the values of the propositions are given in Table \ref{2}. Moreover, all the values of combination are provided in the Table \ref{5}.

\begin{table}[h]\footnotesize
	\centering
	\caption{Evidences given by multiple sensors}
	\begin{spacing}{1.40}
		\begin{tabular}{c c c c c}\hline
			$Evidences$ &\multicolumn{4}{c}{$Values \ \ of \ \ propositions$}\\\hline
			&$\{a\}$&$\{b\}$&$\{c\}$&$\{a,b\}$\\
			$Evidence_{1}$&$0.47$&$0.32$&$0.13$&$0.08$\\  
		    &$\{b\}$&$\{a\}$&$\{c\}$&$\{a,b\}$\\%0.383868 0.271657 0.243264 0.10121
		    $Evidence_{2}$&$0.32$&$0.42$&$0.22$&$0.04$\\ 
		    &$\{c\}$&$\{a\}$&$\{b\}$&$\{a,b\}$\\%0.369174 0.285106 0.244656 0.101063
		    $Evidence_{3}$&$0.28$&$0.44$&$0.12$&$0.16$\\ 
		    &$\{a\}$&$\{b\}$&$\{a,b\}$&$\{c\}$\\%0.365543 0.288928 0.24451 0.101019
		    $Evidence_{4}$&$0.37$&$0.24$&$0.22$&$0.17$\\ %0.386264 0.269028 0.24323 0.101478
			\hline
		\end{tabular}
	\end{spacing}
	%\label{tab:Margin_settings}\K = 0.6683999999999999
	\label{3}
\end{table}

%0.57020848 0.29897384 0.20505194 0.08181296
\begin{table}[h]\footnotesize
	\centering
	\caption{Modified values of the evidence given (Propositions are not ordinal)}
	\begin{spacing}{1.40}
		\begin{tabular}{c c c c c}\hline
			$Evidence$ &\multicolumn{4}{c}{$Values \ \ of \ \ propositions$}\\\hline
			&$\{a\}$&$\{b\}$&$\{c\}$&$\{a,b\}$\\
			$Evidence_{1}$&$0.4932$&$0.2586$&$0.1773$&$0.0707$\\  
			\hline
		\end{tabular}
	\end{spacing}
	%\label{tab:Margin_settings}\K = 0.6683999999999999
	\label{2}
\end{table}

\begin{table}[h]\footnotesize
	\centering
	\caption{Combined values of the evidence given (Propositions are not ordinal)}
	\begin{spacing}{1.40}
		\begin{tabular}{c c c c c}\hline
			$Evidence$ &\multicolumn{4}{c}{$Values \ \ of \ \ propositions$}\\\hline
			&$\{a\}$&$\{b\}$&$\{c\}$&$\{a,b\}$\\
			$Evidence_{1}$&$0.9151$&$0.0691$&$0.0153$&$0.0003$\\  
			\hline
		\end{tabular}
	\end{spacing}
	%\label{tab:Margin_settings}\K = 0.6683999999999999
	\label{5}
\end{table}

It can be easily concluded that the proposition $a$ is the most possible to take place. After checking the values given in the Table \ref{3}, the sequence of proposition $a$ is relatively high and the mass of it is also bigger than others, which means the results of combination are intuitive and reasonable. And for proposition $b$, the sequence and mass of it are lower and smaller than the proposition $a$ but is higher than the rest of the propositions, which is expected to be distributed the second biggest mass to the proposition $b$ and the combined results also indicate that proposition $b$ has a second high degree of probability to take place. With respect to the multiple proposition $\{a, b\}$, it is almost eliminated and the level of the uncertainty of the whole system is reduced to a comparatively extent.

\textbf{Example 2: }
Assume there exists an OFOD which contains four propositions $\{a,b,c,\{a,b\}\}$ and the values of the propositions are listed in the Table \ref{r}. Besides, the extended-dimensional form of the values of the propositions are given in Table \ref{t}. And the results of combination are given in Table \ref{9}.

\begin{table}[h]\footnotesize
	\centering
	\caption{Evidences given by multiple sensors}
	\begin{spacing}{1.40}
		\begin{tabular}{c c c c c}\hline
			$Evidences$ &\multicolumn{4}{c}{$Values \ \ of \ \ propositions$}\\\hline
			&$\{b\}$&$\{c\}$&$\{a\}$&$\{a,b\}$\\
			$Evidence_{1}$&$0.31$&$0.35$&$0.27$&$0.07$\\  %0.373776 0.28039 0.244572 0.101262
			&$\{c\}$&$\{a\}$&$\{b\}$&$\{a,b\}$\\
			$Evidence_{2}$&$0.25$&$0.37$&$0.22$&$0.16$\\ %0.369005 0.285104 0.244647 0.101245
			&$\{b\}$&$\{c\}$&$\{a,b\}$&$\{a\}$\\
			$Evidence_{3}$&$0.34$&$0.32$&$0.26$&$0.08$\\ %0.377835 0.276667 0.244174 0.101324
			&$\{b\}$&$\{c\}$&$\{a,b\}$&$\{a\}$\\
			$Evidence_{4}$&$0.36$&$0.45$&$0.09$&$0.1$\\ %0.368255 0.286538 0.244226 0.100981
			\hline
		\end{tabular}
	\end{spacing}
	%\label{tab:Margin_settings}\K = 0.66051179261041
	\label{r}
\end{table}

%0.17962884 0.4307286 0.40786329 0.10875312
\begin{table}[h]\footnotesize
	\centering
	\caption{Modified values of the evidence given (Propositions are not ordinal)}
	\begin{spacing}{1.40}
		\begin{tabular}{c c c c c}\hline
			$Evidence$ &\multicolumn{4}{c}{$Values \ \ of \ \ propositions$}\\\hline
			&$\{a\}$&$\{b\}$&$\{c\}$&$\{a,b\}$\\
			$Evidence_{1}$&$0.1593$&$0.3821$&$0.3619$&$0.0965$\\  
			\hline
		\end{tabular}
	\end{spacing}
	%\label{tab:Margin_settings}\K = 0.6683999999999999
	\label{t}
\end{table}

\begin{table}[h]\footnotesize
	\centering
	\caption{Combined values of the evidence given (Propositions are not ordinal)}
	\begin{spacing}{1.40}
		\begin{tabular}{c c c c c}\hline
			$Evidence$ &\multicolumn{4}{c}{$Values \ \ of \ \ propositions$}\\\hline
			&$\{a\}$&$\{b\}$&$\{c\}$&$\{a,b\}$\\
			$Evidence_{1}$&$0.0164$&$0.5439$&$0.4373$&$0.0022$\\  
			\hline
		\end{tabular}
	\end{spacing}
	%\label{tab:Margin_settings}\K = 0.6683999999999999
	\label{9}
\end{table}

From the results obtained from the proposed method, it can be summarized that the proposition $b$ owns a very high position in the sequence of proposition and proposition $c$ has relatively bigger values than other propositions which means the values of indicator of the two propositions are supposed to be similar. And for propositions $a$ and $\{a,b\}$, their sequence and values are not placed in an obvious place, so the values combined of them are reduced to indicate both of them can not take place most likely. And the analysis of the results obtained by the proposed method illustrates that all of the values conform to the actual situation and reflect situations of every proposition appropriately.

\subsection{Application of medical diagnosis}
Assume there exists a kind disease diagnosis system and some simple symptoms can be detected. The system can give the probability of a symptom which a patient may possess and the frame of discernment of this system can be defined as $\Theta = \{cough, fever, stomachache, \{cough, fever\}\}$ to describe the actual conditions of the situations of facts. And for one process of detection, the values of different symptoms are given in Table \ref{kkk}. And the results obtained by the proposed method are listed in Table \ref{hhhhh}. In the last, the final judgement of the practical situations are provided in Table \ref{ppppp}.

\begin{table}[h]\footnotesize
	\centering
	\caption{Evidences given by multiple sensors}
	\begin{spacing}{1.40}
		\begin{tabular}{c c c c c}\hline
			$Evidences$ &\multicolumn{4}{c}{$Values \ \ of \ \ propositions$}\\\hline
			&$\{cough\}$&$\{fever\}$&$\{stomachache\}$&$\{cough, fever\}$\\
			$Evidence_{1}$&$0.34$&$0.31$&$0.22$&$0.13$\\  %0.384266 0.270901 0.243464 0.101369
			&$\{fever\}$&$\{cough\}$&$\{cough, fever\}$&$\{stomachache\}$\\
			$Evidence_{2}$&$0.22$&$0.39$&$0.25$&$0.14$\\ %0.372225 0.281911 0.244622 0.101242
			&$\{fever\}$&$\{stomachache\}$&$\{cough\}$&$\{cough, fever\}$\\
			$Evidence_{3}$&$0.25$&$0.28$&$0.33$&$0.14$\\%0.382321 0.272062 0.244073 0.101544
			&$\{stomachache\}$&$\{cough\}$&$\{cough, fever\}$&$\{fever\}$\\
			$Evidence_{4}$&$0.39$&$0.33$&$0.21$&$0.07$\\ %0.384651 0.270613 0.243465 0.101271
			\hline
		\end{tabular}
	\end{spacing}
	%\label{tab:Margin_settings}\K = 0.7303060203833369
	\label{kkk}
\end{table}

%0.41044211 0.26853803 0.29392721 0.13967728
\begin{table}[h]\footnotesize
	\centering
	\caption{Modified values of the evidence given (Propositions are not ordinal)}
	\begin{spacing}{1.40}
		\begin{tabular}{c c c c c}\hline
			$Evidence$ &\multicolumn{4}{c}{$Values \ \ of \ \ propositions$}\\\hline
			&$\{cough\}$&$\{fever\}$&$\{stomachache\}$&$\{cough, fever\}$\\
			$Evidence_{1}$&$0.3689$&$0.2413$&$0.2641$&$0.1255$\\  
			\hline
		\end{tabular}
	\end{spacing}
	%\label{tab:Margin_settings}\K = 0.6683999999999999
	\label{hhhhh}
\end{table}

\begin{table}[h]\footnotesize
	\centering
	\caption{Combined values of the evidence given (Propositions are not ordinal)}
	\begin{spacing}{1.40}
		\begin{tabular}{c c c c c}\hline
			$Evidence$ &\multicolumn{4}{c}{$Values \ \ of \ \ propositions$}\\\hline
		&$\{cough\}$&$\{fever\}$&$\{stomachache\}$&$\{cough, fever\}$\\
			$Evidence_{1}$&$0.6851$&$0.1255$&$0.1801$&$0.0091$\\  
			\hline
		\end{tabular}
	\end{spacing}
	%\label{tab:Margin_settings}\K = 0.6683999999999999
	\label{ppppp}
\end{table}

In this application, two of the evidences indicate that the symptom $fever$ may happen in the first place. However, the probability of this incident to happen is relatively low, which means the two factors may conflict with each other so that the final values of the indicator of the proposition which corresponds to the certain symptom may ba reduced to some relatively low level. Although the symptom $cough$ dose not possess a high position in the sequence of all of the propositions, but the values of it are relatively high which contributes that the final combined values of the proposition $cough$ may have far higher degree than other propositions. And the sequence of the proposition $cough$ is not too non-ideal which also helps it retain a comparatively high probability to take place. And for the symptom $stomachache$, the situation of it is very similar to the one of symptom $cough$, so this values distributed to it lie on the same level of the symptom $cough$. In the last, the values corresponding to the multiple proposition $\{cough, fever\}$ are properly eliminated to reduce the degree of uncertainty of the system given which leads to provide a very clear judgement of the actual situations. All in all, the results obtained by the proposed method are intuitive and reasonable which conform to the practical conditions and may be regarded as a crucial proof for decision makers to make appropriate decisions.

\section{Conclusion}
In this paper, a completely new method to present features of every element contained in an ordinal frame of discernment is proposed utilizing radial basis function and Taylor formula. The extended dimensions of the propositions manifest all of the underlying relations among different evidences, which helps to provide a reliable proof of credibility. The theory proposed in this passage solves the problem that the degree of uncertainty of evidences obtained in the structure of ordinal frame of discernment, which fills the blank field of sequential process of ordered evidences. How to properly combine sequential evidences has been provided a satisfying solution in this paper.
\section*{Acknowledgment}
%The authors greatly appreciate the reviews' suggestions and the editor's encouragement.
 This research was funded by the Chongqing Overseas Scholars Innovation Program (No. cx2018077).

\bibliographystyle{elsarticle-num}
\bibliography{cite}

\begin{thebibliography}{10}
\expandafter\ifx\csname url\endcsname\relax
  \def\url#1{\texttt{#1}}\fi
\expandafter\ifx\csname urlprefix\endcsname\relax\def\urlprefix{URL }\fi
\expandafter\ifx\csname href\endcsname\relax
  \def\href#1#2{#2} \def\path#1{#1}\fi

\bibitem{Deng2019}
X.~Deng, W.~Jiang, A total uncertainty measure for {D} numbers based on belief
  intervals, International Journal of Intelligent Systems 34~(12) (2019)
  3302--3316.
\newblock \href {https://doi.org/10.1002/int.22195}
  {\path{doi:10.1002/int.22195}}.

\bibitem{Kang2019}
B.~Kang, P.~Zhang, Z.~Gao, G.~Chhipi-Shrestha, K.~Hewage, R.~Sadiq,
  Environmental assessment under uncertainty using {Dempster--Shafer} theory
  and {Z}-numbers, Journal of Ambient Intelligence and Humanized Computing
  (2019) DOI: 10.1007/s12652--019--01228--y.

\bibitem{Xiao2019a}
F.~Xiao, {A multiple-criteria decision-making method based on D numbers and
  belief entropy}, International Journal of Fuzzy Systems 21~(4) (2019)
  1144--1153.

\bibitem{book}
G.~Shafer, {A Mathematical Theory of Evidence}, Vol.~1, 1976.
\newblock \href {https://doi.org/10.2307/j.ctv10vm1qb}
  {\path{doi:10.2307/j.ctv10vm1qb}}.

\bibitem{Dempster1967Upper}
A.~P. Dempster, {Upper and Lower Probabilities Induced by a Multi-Valued
  Mapping}, Annals of Mathematical Statistics 38~(2) (1967) 325--339.
\newblock \href {https://doi.org/10.1214/aoms/1177698950}
  {\path{doi:10.1214/aoms/1177698950}}.

\bibitem{li2020newuncertainty}
Y.~Li, H.~Garg, Y.~Deng, {A New Uncertainty Measure of Discrete Z-numbers},
  {International Journal of Fuzzy Systems} {22}~({3}) ({2020}) 760--776.
\newblock \href {https://doi.org/10.1007/s40815-020-00819-8}
  {\path{doi:10.1007/s40815-020-00819-8}}.

\bibitem{Zadeh2011}
L.~A. Zadeh, {A note on Z-numbers}, Information Sciences 181~(14) (2011)
  2923--2932.

\bibitem{tian2020zslf}
Y.~Tian, L.~Liu, X.~Mi, B.~Kang, {ZSLF: A new soft likelihood function based on
  Z-numbers and its application in expert decision system}, IEEE Transactions
  on Fuzzy Systems (2020) DOI: 10.1109/TFUZZ.2020.2997328.

\bibitem{Wang2019}
Y.~Wang, K.~Zhang, Y.~Deng, {Base belief function: an efficient method of
  conflict management}, Journal of Ambient Intelligence and Humanized Computing
  10~(9) (2019) 3427--3437.
\newblock \href {https://doi.org/10.1007/s12652-018-1099-2}
  {\path{doi:10.1007/s12652-018-1099-2}}.

\bibitem{Zhou2020}
Q.~Zhou, H.~Mo, Y.~Deng, A new divergence measure of {Pythagorean} fuzzy sets
  based on belief function and its application in medical diagnosis,
  Mathematics 8~(1) (2020) DOI: 10.3390/math8010142.

\bibitem{Zhang2020}
H.~Zhang, Y.~Deng, Weighted belief function of sensor data fusion in engine
  fault diagnosis, Soft Computing 24~(3) (2020) 2329--2339.

\bibitem{gao2020pseudopascal}
X.~Gao, Y.~Deng, The pseudo-pascal triangle of maximum deng entropy,
  International Journal of Computers Communications \& Control 15~(1) (2020)
  1006.
\newblock \href {https://doi.org/10.15837/3735/ijccc.2020.1.3735}
  {\path{doi:10.15837/3735/ijccc.2020.1.3735}}.

\bibitem{pan2020Probability}
L.~Pan, Y.~Deng, Probability transform based on the ordered weighted averaging
  and entropy difference, International Journal of Computers Communications \&
  Control 15~(4) (2020) 3743.
\newblock \href {https://doi.org/10.15837/ijccc.2020.4.3743}
  {\path{doi:10.15837/ijccc.2020.4.3743}}.

\bibitem{DengeXtropy}
F.~Buono, M.~Longobardi, \href{https://www.mdpi.com/1099-4300/22/5/582}{A dual
  measure of uncertainty: The deng extropy}, Entropy 22~(5) (2020).
\newblock \href {https://doi.org/10.3390/e22050582}
  {\path{doi:10.3390/e22050582}}.
\newline\urlprefix\url{https://www.mdpi.com/1099-4300/22/5/582}

\bibitem{yanhy2020entropy}
H.~Yan, Y.~Deng, An improved belief entropy in evidence theory, IEEE Access
  8~(1) (2020) 57505--57516.
\newblock \href {https://doi.org/10.1109/ACCESS.2020.2982579}
  {\path{doi:10.1109/ACCESS.2020.2982579}}.

\bibitem{Deng2020InformationVolume}
Y.~Deng, Information volume of mass function, International Journal of
  Computers Communications \& Control 15~(6) (2020) 3983.
\newblock \href {https://doi.org/https://doi.org/10.15837/ijccc.2020.6.3983}
  {\path{doi:https://doi.org/10.15837/ijccc.2020.6.3983}}.

\bibitem{deng2021fuzzymembershipfunction}
J.~Deng, Y.~Deng, Information volume of fuzzy membership function,
  International Journal of Computers Communications \& Control 16~(1) (2021)
  4106.
\newblock \href {https://doi.org/https://doi.org/10.15837/ijccc.2021.1.4106}
  {\path{doi:https://doi.org/10.15837/ijccc.2021.1.4106}}.

\bibitem{Deng2020ScienceChina}
Y.~Deng, Uncertainty measure in evidence theory, SCIENCE CHINA Information
  Sciences 64 (2021).
\newblock \href {https://doi.org/10.1007/s11432-020-3006-9}
  {\path{doi:10.1007/s11432-020-3006-9}}.

\bibitem{https://doi.org/10.1002/int.22366}
Y.~He, F.~Xiao,
  \href{https://onlinelibrary.wiley.com/doi/abs/10.1002/int.22366}{Conflicting
  management of evidence combination from the point of improvement of basic
  probability assignment}, International Journal of Intelligent Systems 36~(5)
  (2021) 1914--1942.
\newblock \href
  {http://arxiv.org/abs/https://onlinelibrary.wiley.com/doi/pdf/10.1002/int.22366}
  {\path{arXiv:https://onlinelibrary.wiley.com/doi/pdf/10.1002/int.22366}},
  \href {https://doi.org/https://doi.org/10.1002/int.22366}
  {\path{doi:https://doi.org/10.1002/int.22366}}.
\newline\urlprefix\url{https://onlinelibrary.wiley.com/doi/abs/10.1002/int.22366}

\bibitem{Liu2020Combination}
Z.~Liu, X.~Zhang, J.~Niu, J.~Dezert, Combination of classifiers with different
  frames of discernment based on belief functions, IEEE Transactions on Fuzzy
  Systems (2020) DOI: 10.1109/TFUZZ.2020.2985332.

\bibitem{1111111}
{F.Xiao}, {GIQ: A generalized intelligent quality-based approach for fusing
  multi-source information}, IEEE Transactions on Fuzzy Systems (2020) DOI:
  10.1109/TFUZZ.2020.2991296.

\bibitem{Song2019POWA}
Y.~Song, Y.~Deng, {A new soft likelihood function based on power ordered
  weighted average operator}, International Journal of Intelligent Systems
  34~(11) (2019) 2988--2999.

\bibitem{fei2019evidence}
L.~Fei, Y.~Feng, L.~Liu, Evidence combination using {OWA}-based soft likelihood
  functions, International Journal of Intelligent Systems 34~(9) (2019)
  2269--2290.
\newblock \href {https://doi.org/10.1002/int.22166}
  {\path{doi:10.1002/int.22166}}.

\bibitem{fei2019intuitionistic}
L.~Fei, Y.~Feng, L.~Liu, W.~Mao, On intuitionistic fuzzy decision-making using
  soft likelihood functions, International Journal of Intelligent Systems
  34~(9) (2019) 2225--2242.
\newblock \href {https://doi.org/10.1002/int.22141}
  {\path{doi:10.1002/int.22141}}.

\bibitem{Yager2009}
R.~R. Yager, Weighted maximum entropy owa aggregation with applications to
  decision making under risk, IEEE Transactions on Systems, Man, and
  Cybernetics-Part A: Systems and Humans 39~(3) (2009) 555--564.

\bibitem{Fei2019a}
L.~Fei, Y.~Feng, L.~Liu, Evidence combination using {OWA}-based soft likelihood
  functions, International Journal of Intelligent Systems 34~(9) (2019)
  2269--2290.
\newblock \href {https://doi.org/10.1002/int.22166}
  {\path{doi:10.1002/int.22166}}.

\bibitem{fei2019new}
L.~Fei, H.~Wang, L.~Chen, Y.~Deng, {A new vector valued similarity measure for
  intuitionistic fuzzy sets based on OWA operators}, Iranian Journal of Fuzzy
  Systems 16~(3) (2019) 113--126.

\bibitem{Xiao2019b}
F.~Xiao, Generalization of {Dempster--Shafer} theory: A complex mass function,
  Applied Intelligence 50~(10) (2019) 3266--3275.

\bibitem{Xiao2020b}
F.~Xiao, {CED: A distance for complex mass functions}, IEEE Transactions on
  Neural Networks and Learning Systems (2020) DOI: 10.1109/TNNLS.2020.2984918.

\bibitem{Xiao2020e}
F.~Xiao, Generalized belief function in complex evidence theory, Journal of
  Intelligent \& Fuzzy Systems 38~(4) (2020) 3665--3673.

\bibitem{LiuF2020TFS}
L.~Fan, Y.~Deng, {Determine the number of unknown targets in Open World based
  on Elbow method}, IEEE Transactions on Fuzzy Systems (2020).
\newblock \href {https://doi.org/10.1109/TFUZZ.2020.2966182}
  {\path{doi:10.1109/TFUZZ.2020.2966182}}.

\bibitem{Pan2020}
L.~Pan, Y.~Deng, {An association coefficient of belief function and its
  application in target recognition system}, International Journal of
  Intelligent Systems 35 (2020) 85--104.

\bibitem{Han2018}
Y.~Han, Y.~Deng, An evidential fractal analytic hierarchy process target
  recognition method, Defence Science Journal 68~(4) (2018) 367.

\bibitem{Han2019}
Y.~Han, Y.~Deng, Z.~Cao, C.-T. Lin, An interval-valued {Pythagorean}
  prioritized operator based game theoretical framework with its applications
  in multicriteria group decision making, Neural Computing and Applications
  (2019) DOI: 10.1007/s00521--019--04014--1\href
  {https://doi.org/10.1007/s00521-019-04014-1}
  {\path{doi:10.1007/s00521-019-04014-1}}.

\bibitem{Fei2019b}
L.~Fei, Y.~Feng, L.~Liu, W.~Mao, On intuitionistic fuzzy decision-making using
  soft likelihood functions, International Journal of Intelligent Systems
  34~(9) (2019) 2225--2242.

\bibitem{Song2019a}
Y.~Song, Q.~Fu, Y.-F. Wang, X.~Wang, Divergence-based cross entropy and
  uncertainty measures of {Atanassov's} intuitionistic fuzzy sets with their
  application in decision making, Applied Soft Computing 84 (2019) 105703.

\bibitem{8944285}
F.~{Xiao}, A distance measure for intuitionistic fuzzy sets and its application
  to pattern classification problems, IEEE Transactions on Systems, Man, and
  Cybernetics: Systems (2019) 1--13\href
  {https://doi.org/10.1109/TSMC.2019.2958635}
  {\path{doi:10.1109/TSMC.2019.2958635}}.

\bibitem{Liu2019}
Z.~Liu, Y.~Liu, J.~Dezert, F.~Cuzzolin, Evidence combination based on credal
  belief redistribution for pattern classification, IEEE Transactions on Fuzzy
  Systems (2019) DOI: 10.1109/TFUZZ.2019.2911915.

\bibitem{song2018evidence}
Y.~Song, X.~Wang, W.~Wu, W.~Quan, W.~Huang, Evidence combination based on
  credibility and non-specificity, Pattern Analysis and Applications 21~(1)
  (2018) 167--180.
\newblock \href {https://doi.org/10.1007/s10044-016-0575-6}
  {\path{doi:10.1007/s10044-016-0575-6}}.

\bibitem{feng2019lexicographic}
F.~Feng, M.~Liang, H.~Fujita, R.~R. Yager, X.~Liu, Lexicographic orders of
  intuitionistic fuzzy values and their relationships, Mathematics 7~(2) (2019)
  1--26.

\bibitem{Luo2020negation}
Z.~Luo, Y.~Deng, {A matrix method of basic belief assignment's negation in
  Dempster-Shafer theory}, IEEE Transactions on Fuzzy Systems 28~(9) (2020)
  2270--2276.

\bibitem{Fei2019}
L.~Fei, J.~Xia, Y.~Feng, L.~Liu, An {ELECTRE}-based multiple criteria decision
  making method for supplier selection using {Dempster-Shafer} theory, IEEE
  Access 7 (2019) 84701--84716.
\newblock \href {https://doi.org/10.1109/ACCESS.2019.2924945}
  {\path{doi:10.1109/ACCESS.2019.2924945}}.

\bibitem{Deng2019b}
X.~Deng, W.~Jiang, Z.~Wang, {Zero-sum polymatrix games with link uncertainty: A
  Dempster-Shafer theory solution}, Applied Mathematics and Computation 340
  (2019) 101--112.

\bibitem{yager1987dempster}
R.~R. Yager, {On the Dempster-Shafer framework and new combination rules},
  Information sciences 41~(2) (1987) 93--137.

\end{thebibliography}
\end{document}